\documentclass[10pt,twocolumn,letterpaper]{article}
\usepackage{cvpr}
\usepackage{times}
\usepackage{graphicx}
\usepackage{amsmath}
\usepackage{amssymb}

\usepackage{comment}
\usepackage{placeins}

\usepackage{algpseudocode,algorithm,algorithmicx}
\usepackage{tikz}
\usepackage{eucal}
\usepackage{enumitem}
\usepackage{color}

\usepackage[pagebackref=true,breaklinks=true,letterpaper=true,colorlinks,bookmarks=false]{hyperref} 

\cvprfinalcopy 

\def\cvprPaperID{0035} 

\ifcvprfinal\fi

\newcommand{\diag}{\mathrm{diag}}
\newcommand{\trans}{^\mathsf{T}}

\newcommand{\myfigref}[1]{Fig.~\ref{#1}}
\newcommand{\myeqref}[1]{Eq.~\eqref{#1}}

\newcommand{\imsizeX}{X} 
\newcommand{\imsizeY}{Y} 
\newcommand{\patchsize}{M} 
\newcommand{\dictsize}{K} 
\newcommand{\biadjacency}{\mathbf{B}} 
\newcommand{\imtodict}{\mathbf{T}_1} 
\newcommand{\dicttoim}{\mathbf{T}_2} 

\begin{document}

\title{Content-based Propagation of User Markings for\\Interactive Segmentation of Patterned Images}
\author{Vedrana A. Dahl,  Monica J. Emerson, Camilla H. Trinderup, and Anders B. Dahl\\
Department of Applied Mathematics and Computer Science, Technical University of Denmark\\
Richard Petersens Plads, Building 324, 2800 Kgs.~Lyngby, Denmark\\
{\tt\small {vand,monj,ctri,abda}@dtu.dk}}

\maketitle

\begin{abstract}
Efficient and easy segmentation of images and volumes is of great practical importance. Segmentation problems that motivate our approach originate from microscopy imaging commonly used in materials science, medicine, and biology. We formulate image segmentation as a probabilistic pixel classification problem, and we apply segmentation as a step towards characterising image content. Our method allows the user to define structures of interest by interactively marking a subset of pixels. Thanks to the real-time feedback, the user can place new markings strategically, depending on the current outcome. The final pixel classification may be obtained from a very modest user input. An important ingredient of our method is a graph that encodes image content. This graph is built in an unsupervised manner during initialisation and is based on clustering of image features. Since we combine a limited amount of user-labelled data with the clustering information obtained from the unlabelled parts of the image, our method fits in the general framework of semi-supervised learning. We demonstrate how this can be a very efficient approach to segmentation through pixel classification.
\end{abstract}

\section{Introduction}
\label{intro}
In this paper, we propose an interactive method for probabilistic classification of pixels, which can be used for segmentation of 2D and 3D images. Our approach is especially advantageous for detecting patterns, a situation regularly occurring in microscopy of materials and medical samples. Such images often show a collection of objects which are to be separated from the background. For example consider segmenting individual facets of a bee eye shown in Fig~\ref{fig:eyes}.

When segmenting images showing a collection of similar objects, an established strategy involves extensive modelling of the appearance of the objects, usually leading to a highly specialised method. Another common strategy is to learn the appearance of the objects from a large amount of prelabelled data, often with high computational requirements during the training phase. Here we aim for a general method that requires limited computation, as well as modest user-labelling.

Our method fits into the framework of semi-supervised learning, combining two ingredients: a model for image content created in an unsupervised manner from the image features, and a modest input from the user. When a user marks a structure in the image as belonging to a class, our method propagates the marks to similar structures in the rest of the image. The output is a layered image which at every pixel position contains the probabilities of belonging to each of the defined classes. We call this output \emph{pixelwise probabilities} of belonging to segmentation classes. From pixelwise probabilities, the segmentation is readily obtained by selecting the most probable class for each pixel. The method is highly flexible and captures the features which are of interest to the user; an example with various image features is  shown in Fig.~\ref{fig:bag}. Our approach allows easy segmentation of complex structures, that would otherwise require the development of algorithms targeted at specific problems.

An important property of our model is real-time feedback, allowing the user to place new markings strategically, depending on the current result. For this to work without delay, the segmentation must be updated very fast. Our method relies on an efficient update of the parameters used for pixel classification, and an equally efficient update of the classification results. With results shown promptly, the user can continue adding marks until the desired outcome is learned by the algorithm. Having learned the desired outcome, the classification model can be applied to other images of the same type in an unsupervised manner, that is, without additional user input. 

Our prototype implementation, including a graphical user interface, is in Matlab and C++. The code is available at \href{https://github.com/vedranaa/InSegt}{https://github.com/vedranaa/InSegt}.

\newlength{\vsep}\setlength{\vsep}{0.01\linewidth}
\newlength{\sub}\setlength{\sub}{0.495\linewidth}

\begin{figure}
\centering
  \includegraphics[width=\sub]{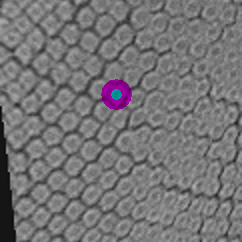}\hspace{\vsep}%
  \includegraphics[width=\sub]{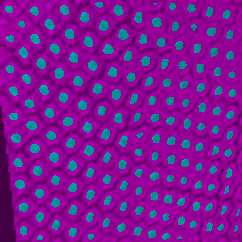}%
\caption{Detecting individual facets of a bee eye using our interactive pattern-based segmentation method. On the \emph{left} input image and a very small subset of pixels manually marked as either being close to a facet centre (cyan) or not being close to a facet centre (magenta). On the \emph{right}, the manual labelling has been propagated to the whole image and the result is obtained by selecting the most probable class for each pixel.}
\label{fig:eyes}
\end{figure}

\def \newsufix {_displayed}

\setlength{\sub}{0.1377\textwidth}
\begin{figure*}%
\centering%
  \includegraphics[width=\sub]{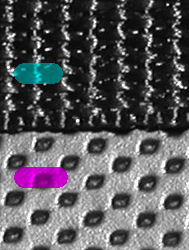}\hspace{\vsep}%
  \includegraphics[width=\sub]{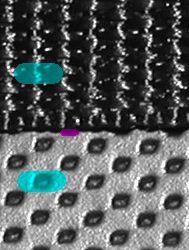}\hspace{\vsep}%
  \includegraphics[width=\sub]{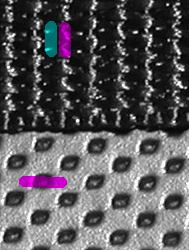}\hspace{\vsep}%
  \includegraphics[width=\sub]{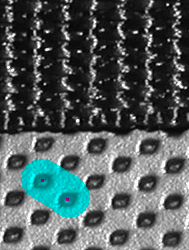}\hspace{\vsep}%
  \includegraphics[width=\sub]{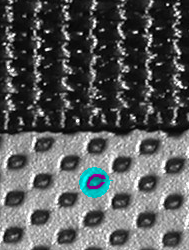}\hspace{\vsep}%
  \includegraphics[width=\sub]{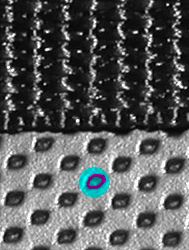}\hspace{\vsep}%
  \includegraphics[width=\sub]{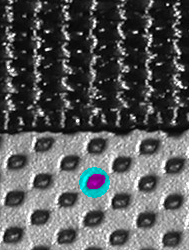}%
  \\\vspace{\vsep}%
  \includegraphics[width=\sub]{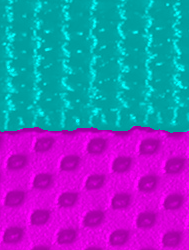}\hspace{\vsep}%
  \includegraphics[width=\sub]{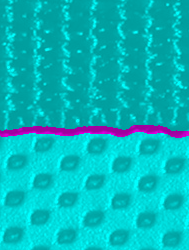}\hspace{\vsep}%
  \includegraphics[width=\sub]{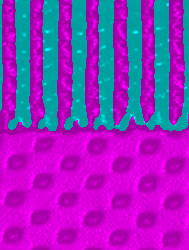}\hspace{\vsep}%
  \includegraphics[width=\sub]{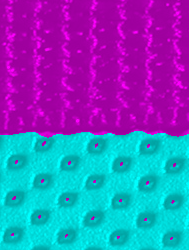}\hspace{\vsep}%
  \includegraphics[width=\sub]{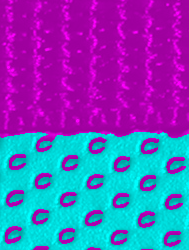}\hspace{\vsep}%
  \includegraphics[width=\sub]{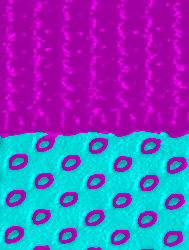}\hspace{\vsep}%
  \includegraphics[width=\sub]{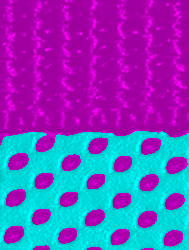}%
\caption{An example demonstrating the flexibility of our method. In the \emph{top} row, different features of interest marked by the user in two classes (cyan and magenta). The user input is propagated between image pixels using the same model for image content, but results in different segmentations, shown in the \emph{bottom} row.}
\label{fig:bag}
\end{figure*}

\subsection{Related work}
Benefits of user input with real-time feedback have been recognised in image segmentation. A comprehensive summary of interactive approaches can be found in \cite{boykov2014computer}. Here, we review some important advances to place our method in the existing framework, and to explain how our method differs from the current trends in interactive segmentation. 

Early interactive techniques for segmentation of highly complex images include intelligent scissors \cite{mortensen1995intelligent} or live wire \cite{falcao1998user}, where the user cuts out an object by placing markers along its boundary. 
These algorithms are computationally cheap but require a lot of user effort to obtain a segmentation. Less user input is required when using interactive graph cuts \cite{boykov2001interactive,boykov2006graph}, which often give very impressive results with only a few seeds provided by the user. 
In the GrabCut method \cite{rother2004grabcut} the user provides a bounding rectangle, often leading to very precise foreground-background separation. 
Extensions of GrabCut include shape priors \cite{price2010geodesic} and and improvement to graph cut energy representation \cite{tang2013grabcut}. An alternative to combinatorial graph-based solutions is the use of a curve to represent the segmentation boundaries. Such interactive active contours often minimise an energy functional in a variational framework \cite{unger2008tvseg,santner2011interactive}.


Common to the described methods is the focus on segmenting relatively large foreground objects, which justifies using regularisation on the length or the curvature of the segmentation boundary. In some applications, it is, however, not possible to use a strong regulariser. For example, when segmenting the bee eyes shown in \myfigref{fig:eyes}, regularisation could remove or merge small regions. 

The need for segmenting a number of small objects is often seen in areas like microscopy for life science or materials science. The appearance of such images can vary significantly, with texture as well as intensity carrying information that is useful for obtaining the desired segmentation. A specialist would use such clues to distinguish amongst structures, but automating the segmentation task typically requires highly sophisticated and problem-adapted methods. While there are situations that justify the development of a specialised method, in many cases a reasonable result with modest interactive effort would be preferred. 

When segmenting small image structures, e.g.\ cells, a well-suited approach is classification of pixels. This is the basis for the ilastik segmentation tool \cite{berg2019ilastik,sommer2011ilastik}, which employs a random forest classifier \cite{breiman2001random} trained on image features including colour, edges, orientation and texture. The features are computed from the image before starting the interactive labelling of image structures, while parameters of the random forest classifier are learned from the manual labelling. When a user updates the labels to improve the segmentation, the parameters of the classifier need to be re-learned, which is computationally costly and causes a noticeable delay in the feedback. 
Another specialised tool for segmentation of microscopy images is the trainable Weka segmentation \cite{arganda2017trainable} (a part of the Fiji \cite{schindelin2012fiji} distribution of ImageJ) which utilises a data mining and machine learning toolkit for solving pixel classification problems. A user can choose from a variety of image features and interactively re-train the classifier.

Frameworks using neural networks are increasingly popular in pixel classification, and often yield impressive results \cite{lecun2015deep}. A neural network operates on features extracted locally from the image. This input is fed through a series of multidimensional linear functions, with a non-linear activation between them, ending up in a probabilistic output. The weights of the linear functions need to be trained by optimising the performance on the usually large set of prelabelled data. This provides extreme flexibility to the method and, provided adequate training, neural networks may solve pixel classification problems as accurately as specialists. However, neural networks are dependent on large training sets and require computationally costly training, which makes them less convenient for the task of segmenting a small set of images.

Our approach shares some similarities with neural networks. We also feed the input through linear functions with non-linear steps in between. However, we use the features extracted from the image to construct the linear functions in a preprocessing step. The functions are then kept fixed, while they operate on the interactively provided user input, resulting in a probabilistic output. Due to the fixed linear functions, our method is not as adaptable as neural networks. For example, our approach is less fit for semantic segmentation of photographs. Nevertheless, we achieve excellent results when segmenting patterned images, without requiring a large set of labelled data and without performing a costly optimisation during interactive update.

The foundation of our method is a linear operator encoding image content using  image--dictionary relationship. Similar approach, without the interactive update, has been used for evolving deformable models \cite{dahl2014dictionary,dahl2015dictionary,dahl2017probabilistic}, quantifying composite materials \cite{emerson2018statistical,emerson2018quantifying} and measuring retinal microvasculature \cite{engberg2020automated}. In this work, we use the image--dictionary relationship to propagate the brush strokes provided by the users.

\section{Method}
\label{sec:method}

Our method combines two sources of information, the structure in the image and the user-provided partial labelling. The structure in the image is captured in the preprocessing step, namely clustering, which we describe in \ref{subsec:preprocessing}. After that, in \ref{subsec:transformation}, we explain how clustering is used for transforming the user-provided partial labelling into pixelwise probabilities of belonging to each of the classes. The interactive update, covered in \ref{subsec:update}, is obtained by immediately displaying the result of the transformation and allowing the user to repeatedly improve the partial labelling.

Postprocessing choices, covered in \ref{subsec:postprocessing}, are concerned with the outputs of the interactive update. The most obvious output is a \emph{probability image}. While probability image can give the image segmentation, other postprocessing methods may be utilised as well. For the second output, which we call \emph{dictionary probabilities}, the user-provided partial labellings are propagated to the clusters constructed in the preprocessing step. This encodes the learned information about the structures in the image and can be used for subsequent automatic processing of similar images.

Our method comes in a range of flavours. In this section, we explain only the simplest variant, the other possibilities are covered in Sec.~\ref{sec:implementation}.

\emph{Notation.} Throughout the paper we consider an image $I$ defined on an $\imsizeX $-by-$\imsizeY$ image grid with pixel values in either grayscale or RGB colour space. During the interactive part, the user will be placing marks in the image grid, to indicate the pixels which belong to one of the $C$ segmentation classes. We chose to represent this user-provided information with a layered \emph{label image} $L$, where $L(x,y,c) = 1$ if the user indicated that pixel $(x,y)$ belongs to class $c$, and 0 otherwise.

\subsection{Clustering image patches}
\label{subsec:preprocessing}
The aim of preprocessing is to find the structures in the image without considering the user-provided labels. In the framework of semi-supervised learning, a cluster assumption states that, if points are in the same cluster, they are likely to be of the same class -- which does not imply that each class forms a single cluster \cite{chapelle2009semi}. For our purpose, we assume that image features tend to form discrete clusters and that image features in the same cluster are more likely to share a class. However, we do not assume that each class is represented by only one cluster, so we will need many more clusters than classes. Therefore, we create a multitude of clusters to capture the variety of features present in the image.

In Sec.~\ref{sec:implementation} we will explain the implementation details and some more advanced ways of accomplishing clustering. In this section, we outline the basic approach, which operates on intensity patches. For this case, only two parameters are required: the number of clusters $\dictsize$ and the size of the patches $\patchsize$. The number of clusters should be large, measured in hundreds or thousands, and is roughly reflecting the variability in the image. The size of the patches should reflect the scale of the distinctive image features and could, for example, be 9 pixels. For simplicity, we always assume that the size of the image patches $\patchsize$ is odd and patches are centred around the central pixel.

We extract patches of size $\patchsize$-by-$\patchsize$ from the image $I$, treat each patch as a vector containing the pixel intensities and group those vectors into $\dictsize$ clusters, e.g.\ using $k$-means clustering based on Euclidean distance. The resulting collection of cluster centres represents the content of the image. As these basic elements are inferred by grouping features from the image, we call the collection of $\dictsize$ cluster centres an \emph{intensity dictionary}, and each of its elements (each cluster centre) is denoted \emph{dictionary patch}. Every image pixel $(x,y)$ in the centre of an $\patchsize $-by-$\patchsize$ image patch is, by means of clustering, uniquely assigned to one cluster. We represent this using an \emph{assignment image} $A$. For boundary pixels we define $A(x,y) = 0$.

\subsection{Relation between image and dictionary} 
\label{subsec:transformation}
According to the cluster assumption, image patches assigned to the same dictionary patch are more likely to belong to the same class. Unique for our method is that we use this assumption on a pixel level, and not on a patch level. That is, if two image patches are assigned to the same dictionary patch, their \emph{corresponding} pixels (i.e.\ the pixels at the same position in the patch) are more likely to belong to the same class. In other words, for every dictionary patch, there is a certain (unknown) classification of its individual pixels, which all assigned patches are likely to share.

To exploit this assumption, we define a binary relation between corresponding pixels assigned to the same dictionary pixel. For example, a central pixel of an image patch assigned to a certain dictionary patch relates to central pixels of all other patches assigned to the same dictionary patch. Likewise, the pixel directly above the central pixel relates to corresponding pixels in other patches, and a similar relation extends to all positions in a patch. This results in $M^2K$ cliques of pixels, one for every pixel in the intensity dictionary. Due to the overlap between image patches, every non-boundary pixel belongs to $M^2$ different cliques.

The central part of our method is concerned with transforming a user-provided partial labelling to pixelwise probabilities. The transformation matrix we use has a very simple decomposition, which makes our method efficient and allows for immediate feedback to the user. The construction of the transformation matrix is therefore fundamental for our method. However, describing how this matrix is constructed provides little intuition about our method, so we start by motivating our approach.

As covered previously, the assignment image $A$, obtained in an unsupervised manner, contains information on clusters of structures in the image $I$. At the same time, image $I$ is accompanied by the user-provided partial labelling $L$. To combine the two sources of information, we create a dictionary of labels to accompany our intensity dictionary. For each dictionary patch $k\in\{1,\dots,K\}$ we use $A$ to identify the locations of all image patches assigned to it. At those locations in the image grid we extract corresponding patches but from the labelling image $L$. For the set of related labelling patches we compute a pixelwise average for every layer. As a result, every $M$-by-$M$ dictionary patch now has a corresponding $M$-by-$M$ labelling representation consisting of $C$ layers.

When the image is fully labelled, the label image $L$ sums to one in every pixel, as only one out of $C$ classes has a label of 1. Consequently, the labelling representation of every dictionary patch also sums to one in every pixel. However, due to the pixelwise averaging, the values of this representation are not binary, they instead encode the normalised frequency of a dictionary pixel being labelled as belonging to class $c$ in the current labelling image. For this reason, we think of this labelling representation as of pixelwise probabilities of belonging to class $c$, and we call them \emph{dictionary probabilities}.

Dictionary probabilities can now be pasted back into an $\imsizeX $-by-$\imsizeY$ image grid, again using the location information from $A$, and again averaged in every pixel. This results in an $\imsizeX $-by-$\imsizeY$ probability image $P$ consisting of $C$ layers, where $P$ is a diffused version of $L$. In other words, we use the self-similarity information encoded by $A$ to propagate the user-provided markings from $L$ onto the rest of the image. 

In light of this motivation, now we turn to explain the construction of the transformation matrices used for efficient computation of dictionary probabilities and image probabilities. Fundamental for this transformation is the relation between the $\imsizeX $-by-$\imsizeY$ image grid and the $\patchsize$-by-$ \patchsize$-by-$\dictsize$ dictionary grid. This relation will be encoded using an $n$-by-$m$ biadjacency matrix $\mathbf{B}$, where $n = XY$ and $m = M^2K$. For this purpose, we need a linear (single) index for the pixels in the image and the pixels in the dictionary grid.

The linear index of an image pixel $(x,y)$ is
\begin{equation}
i = x + (y-1)\imsizeX\,.\label{eq:image_linear_index}
\end{equation}

As for the dictionary grid, we use $(0,0,k)$ for the central pixel of the $k$-th dictionary element, and coordinates of other pixels in the patch are defined in terms of within-patch displacements $\Delta x$ and $\Delta y$, both from $\{-s,\dots,0,\dots,s\}$ with $s = (M-1)/2$. A dictionary pixel at coordinates $(\Delta x,\Delta y,k)$  has a linear index
\begin{equation}
j = (\Delta x+s) + (\Delta y+s) \patchsize + (k-1)\patchsize^2\,.\label{eq:dict_linear_index}
\end{equation}

Each assignment of an image patch centered around $(x,y)$ to a $k$-th dictionary patch centered around $(0,0,k)$ induces a relation between the $\patchsize^2$ image pixels and the $\patchsize^2$ dictionary pixels, see \myfigref{fig:ID_relation_smaller}. 
Using $\sim$ for denoting a relation between image pixels and dictionary pixels gives
\begin{equation}
A(x,y)=k \; \Rightarrow \: \begin{array}{l} (x{+}\Delta x,y{+}\Delta y)\sim(\Delta x,\Delta y,k),\\ \text{for all }\Delta x \text{ and }\Delta y\end{array} \,.\label{eq:relation}
\end{equation}

\begin{figure}
\begin{center}
\definecolor{mycolor}{rgb}{0.4,0.2,1} 
\begin{tikzpicture}[scale=\linewidth/10cm]
\clip (-0.6,-0.7) rectangle (9.4,3);
\draw [<->] (-0.2,0.5) node[left] {\small$x$} --(-0.2,-0.2)--(0.5,-0.2) node[below] {\small$y$};
\draw [<->] (5.8,0.5) node[left] {\small$\Delta x$} --(5.8,-0.2)--(6.3,-0.2) node[below] {\small$\Delta y$};
\draw [->] (7.7,-0.2) --(8.3,0.2) node[below] {\small$k$};

\fill [black!5!white] (0,0) rectangle (4.5,3); 
\fill [black!20!white] (1,1) rectangle (1.5,1.5); 
\draw [black,thin,step=0.5] (0,0) grid (4.5,3); 
\draw[black!80!white, very thick] (0.5,0.5) rectangle (2,2); 
\foreach \dx/\dy in {{7.8/1.2},{7.2/0.8},{6.6/0.4},{6/0}}
{
    \fill [black!5!white,xshift=\dx cm,yshift=\dy cm] (0,0) rectangle (1.5,1.5); 
    \draw [black,thin,step=0.5,xshift=\dx cm,yshift=\dy cm] (0,0) grid (1.5,1.5); 
}
\draw[black!80!white, very thick] (6,0) rectangle (7.5,1.5); 
\foreach \y in {2,3,4}\foreach \x in {3,4,5}
{
    \node[circle,inner sep=0pt,text width=0.1cm,mycolor,fill] (I\x\y) at (0.5*\x-0.75,0.5*\y-0.25){};
    \node[circle,inner sep=0pt,text width=0.1cm,mycolor,fill] (A\x\y) at (5.5+0.5*\x-0.25-0.5,0+0.5*\y-0.25-0.5) {};
    \draw[mycolor,thick] (I\x\y) to [bend left=30] (A\x\y);
}
\end{tikzpicture}
\end{center}
\caption{\label{fig:relations} A subset of relations between a $9\times 6$ image and a $3\times3\times4$ dictionary caused by the framed patch centered around the pixel shaded darker being assigned to the first dictionary patch.
\label{fig:ID_relation_smaller}}
\end{figure}

Since image patches are overlapping, every non-boundary image pixel relates to $\patchsize^2$ dictionary pixels. Image pixels in a boundary relate to less than $\patchsize^2$ dictionary pixels, and the four corner pixels relate to only one dictionary pixel. In total there are $(\imsizeX-2s)(\imsizeY-2s)\patchsize^2$ relations between the image pixels and the dictionary pixels. 

We represent the relations between $n$ image pixels and $m$ dictionary pixels using an $n$-by-$m$ biadjacency matrix $\biadjacency$, with elements
\begin{equation}
b_{ij} = \left\{\begin{array}{l l}
    1 & \quad i\sim j \\ 0 & \quad \mathrm{otherwise}
  \end{array}\right. \quad ,
\end{equation}
where $i$ and $j$ are linear indices of an image pixel 
and a dictionary pixel. 
The algorithm for constructing $\biadjacency$ is summarised in Alg.~\ref{alg:biadjacency}.

\begin{algorithm}
  \caption{Construction of $\biadjacency$}
  \begin{algorithmic}[1]
    \State Initiate $\biadjacency$ as an $n$-by-$m$ matrix with $b_{ij}=0$
      \For{an non-boundary pixel $(x,y)$}
        \State Retrieve pixel assignment $k = A(x,y)$
        \For{within-patch displacement $(\Delta x,\Delta y)$}
         \State compute $i$ for $(x+\Delta x,y + \Delta y)$ using \myeqref{eq:image_linear_index}
         \State compute $j$ for $(\Delta x,\Delta y,k)$ using \myeqref{eq:dict_linear_index}
         \State assign $b_{ij} = 1$ 
        \EndFor
      \EndFor
  \end{algorithmic}
  \label{alg:biadjacency}
\end{algorithm}

The biadjacency matrix $\biadjacency$ defines the linear mapping used to propagate the information from the image to the dictionary and vice versa. Consider a quantity defined on the image grid (e.g. user-provided markings indicating pixels which belong to class 1) arranged into a length $n$ vector $\mathbf{v}$ such that the $i$-th element contains the value of the $i$-th image pixel. Propagating these values to the dictionary is carried out by calculating a length $m$ vector
\begin{equation}
\mathbf{d} = \diag(\biadjacency\trans\mathbf{1}_{n\times1})^{-1}\biadjacency\trans \mathbf{v}\,, \label{eq:L_to_D}
\end{equation}
where $\mathbf{1}$ denotes a column vector of ones, while $\diag(\cdot)$ denotes a diagonal matrix with the diagonal defined by the argument. The $j$-th element of $\mathbf{d}$ contains the value of the $j$-th dictionary pixel computed by averaging the values of the related image pixels. The summation is accomplished by multiplying with $\biadjacency\trans$ while the diagonal matrix accomplishes the division with the total number of related pixels.

For this reason we define the $m$-by-$n$ transformation matrix for mapping from the image to the dictionary as
\begin{equation}
\imtodict = \diag(\biadjacency\trans\mathbf{1}_{n\times1})^{-1}\biadjacency\trans \,. \label{eq:T1}
\end{equation}
Similarly, mapping from the dictionary to the image is given by the $n$-by-$m$ matrix
\begin{equation}
\dicttoim = \diag(\biadjacency\mathbf{1}_{m\times1})^{-1}\biadjacency\,. \label{eq:T2}
\end{equation}

Those two transformation matrices are fundamental for our method. The propagation of user-provided markings (as described in the motivational paragraphs) is computed as
\begin{equation}
\mathbf{P} = \dicttoim \imtodict \mathbf{L} \,, \label{eq:L2P}
\end{equation}
where $\mathbf{L}$ is the user-provided labelling $L$ arranged in a $n$-by-$C$ matrix, while the resulting $n$-by-$C$ matrix $\mathbf{P}$ needs to be arranged back into a layered image $P$.

\subsection{Interactive update}
\label{subsec:update}

When equipping our method with the user-provided interactive update, we run into choices with regards to: i) how we treat unlabelled pixels, ii) the number of applied diffusion steps, and the way of treating intermediate results between the steps, and iii) the possibility of changing the number of segmentation classes. After testing many types of interactive updates, we kept three main versions. In all versions the number of classes $C$ is chosen during initialisation and kept fixed during the update.

How we handle pixels that have not been labelled by the user is also common to all versions. Such pixels are initially assigned equal probability of belonging to each class. As a result, before the user places the first label, all probabilities are equal and no segmentation is possible.

The user starts the interaction by choosing a pencil corresponding to one of the $C$ classes applies markings to some pixels. The partial labelling information is immediately transformed to the probability image and shown to the user as an image segmentation, with every pixel placed in the class with the highest probability. After the first pencil stroke, only one class will have values larger than $\frac{1}{C}$ in the label image $L$, and the same applies for the probability image $P$ computed using \eqref{eq:L2P}. Thus, at first, many pixels will belong to the first marked class and no pixels will be assigned to the classes that have not been marked yet. As the user adds markings for the other classes, those will appear in probability image $P$.

Thanks to the real-time feedback, the user can quickly improve the result by placing markings in misclassified regions (the regions that have been incorrectly classified). With many unlabelled pixels in $L$, the image $P$ will typically have many values that only differ slightly from $\frac{1}{C}$. Those small deviations carry the information needed for inferring the class of the unlabelled pixels.

As for the number of applied diffusion steps, we use either one or two. When using two diffusion steps, instead of continuing to diffuse the (already diffused) probability image, we can apply additional non-linear operations between the two diffusions. Very good results are obtained if we apply \emph{binarisation} of the labels between the two diffusion steps. For binarisation, we identify the class of the highest probability for each pixel, and apply $\{0,1\}$ labelling. If there are pixels with no clear probability maximum, we let them retain their unresolved labels. Consequently, for the second iteration of the diffusion, many pixels act as labelled, and this improves the quality of the result.

The options for the two-step diffusion and binarisation are implemented in our segmentation tool, such that the user can quickly switch between the variants of the method and decide which one yields the best results for the data at hand. Likewise, the user can quickly determine whether the quality of the results is sufficient or additional markings should be placed.

The user can inspect the output of the classification displayed as a final segmentation based on the resulting probability image. Alternatively, there is an option for inspecting the $C$ probability images, which often gives a better insight into the quality of the result. 

\subsection{Postprocessing}
\label{subsec:postprocessing}
Our approach allows for various postprocessing options, which may be grouped into two postprocessing strategies. One strategy involves processing the probability image to obtain the segmentation or detection of interesting features from the probability image. These operations are application-driven and examples are illustrated in Sec.~\ref{sec:results}.

The second strategy involves reusing the information stored in the dictionary and the associated dictionary probabilities. The linear transformation \eqref{eq:L2P}, which is core to our method, first transforms the user-provided markings from $L$ to the dictionary space (using matrix $\imtodict$) and then back to the image space (using matrix $\dicttoim$). Consider only the first product
\[\mathbf{D} = \imtodict\mathbf{L}\,.\]
This is an $m$-by-$C$ matrix containing the pixelwise probabilities of the dictionary pixels (i.e. the dictionary probabilities) which can be useful for processing a previously unseen image similar to $I$.

Processing a new image $\hat{I}$ requires extracting all $M$-by-$M$ patches for every pixel of $\hat{I}$ and assigning those patches to the \emph{existing} dictionary, i.e.\ the dictionary created using patches from $I$. Just like before, this assignment defines an image-to-dictionary and we can compute the two associated transformation matrices. Here we are interested in the dictionary-to-image transformation $\hat{\mathbf{T}}_2$. To compute the probability image corresponding to the unlabelled image $\hat{I}$ we therefore need to compute
\[\hat{\mathbf{P}} = \hat{\mathbf{T}}_2\mathbf{D}\,.\]
and rearrange the result into $\hat{P}$.

This way of using our method fits into the framework of supervised learning. The original image $I$ and the computed labelling $L$ can in this context be seen as a (labelled) training set (ignoring the fact that the labelling is computed in a semi-supervised way). Our method is then capable of producing the probability image $\hat{P}$ for the new, unlabelled image $\hat{I}$. The approach will work as long as the initial clustering captures the features present in $\hat{I}$, which holds for similar images.

\section{Implementation details}
\label{sec:implementation}

When developing our framework, we made a number of implementational choices governed by the performance of our method. First, our method is rather robust to the \emph{quality} of the clustering while preprocessing, so using an approximate clustering will generally not deteriorate the output. 
We therefore focus on efficiency when building the dictionary and use a $k$-means tree \cite{nister2006scalable}, built from consecutive $k$-means clusterings. In this implementation, the size of the dictionary is defined in terms of the branching factor $b$ and the number of layers $t$. Since each node in the tree makes up a dictionary element, the total number of dictionary elements is given by $K = \frac{b^{t+1}-1}{b-1}$.

Our experience is that good performance is obtained also without running the $k$-means until convergence for each tree layer, and therefore a fixed number of iterations is chosen, e.g.~10 iterations. Furthermore, to limit the computational burden and memory usage, we extract only a subset of $\patchsize$-by-$\patchsize$ patches from the image when building the dictionary.

As for producing $A$ given the clustering represented by a $k$-means tree, the patch vector is compared with the nodes in the first layer to find the match. The patch vector is then compared to the children of this node, and the most similar node is again chosen. This process is repeated until a leaf node or an empty node is reached. The patch vector is assigned to the most similar node along this path.

Second, the features used for clustering need to reflect the distinction in the appearance of the classes we want to separate. For many types of images, an intensity-based approach as sketched in Sec.~\ref{sec:method} will perform well. However, in challenging cases, more elaborate image features might provide better results. Some of the results we show in Sec.~\ref{sec:results} are based on SIFT \cite{lowe2004distinctive}, but other features can also be incorporated in our method. The approach is as follows. 

Image features represented by vectors are extracted from all pixel positions in the image and clustered in $K$ clusters. For speed, it often suffices to consider only a subset of pixels for clustering, as long as we capture the variability in the image. Every position $(x,y)$ from the image grid can now be uniquely assigned to one of the $k$ clusters -- the cluster that is closest to the feature vector extracted at $(x,y)$. This results in an assignment image $A$. The only additional information we need for building the transformation matrices is a value $M$, which earlier represented the size of the extracted image patches. The value $M$ now determines the size of the overlap when linking the image to the dictionary. While we now freely chose $M$, it is reasonable to use a value that corresponds to the size of the extracted features.

\section{Results}
\label{sec:results}

\setlength{\sub}{0.495\linewidth}
\begin{figure}[t]
\centering
  \includegraphics[width=\sub]{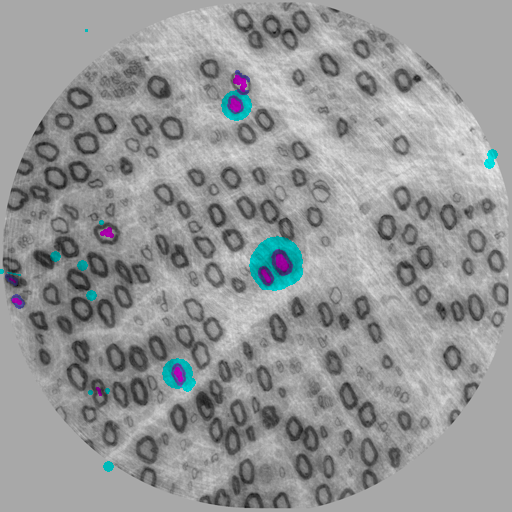}\hspace{\vsep}%
  \includegraphics[width=\sub]{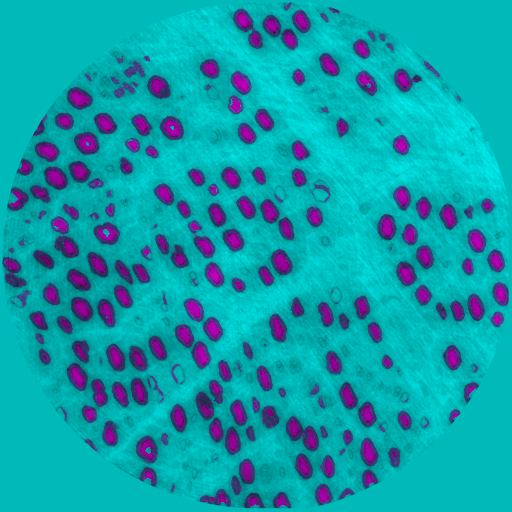}\\\vspace{\vsep}
  \includegraphics[width=\sub]{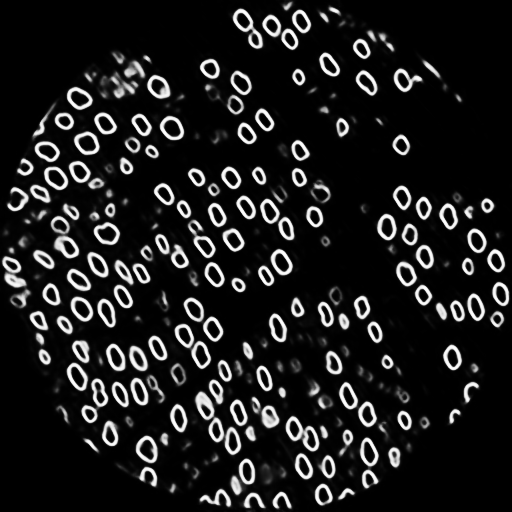}\hspace{\vsep}%
  \includegraphics[width=\sub]{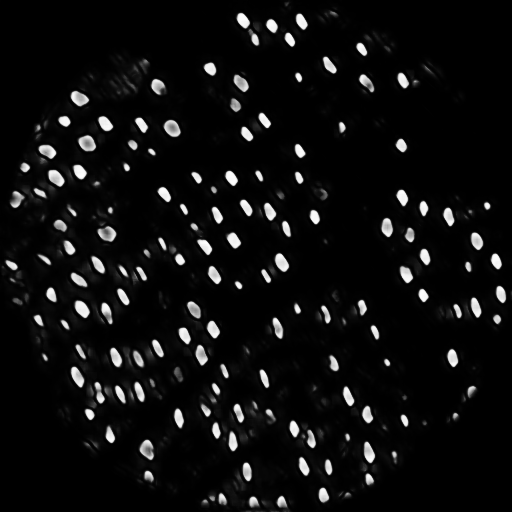}\\\vspace{\vsep}%
  \includegraphics[width=\sub,trim={5.9cm 2.3cm 5.9cm 2.3cm},clip]{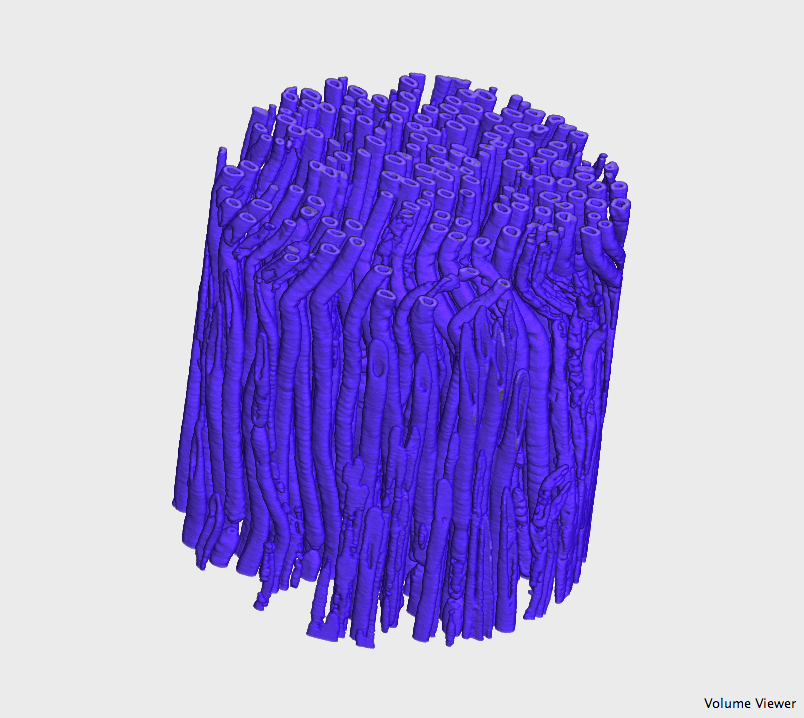}\hspace{\vsep}%
  \includegraphics[width=\sub,trim={5.9cm 2.3cm 5.9cm 2.3cm},clip]{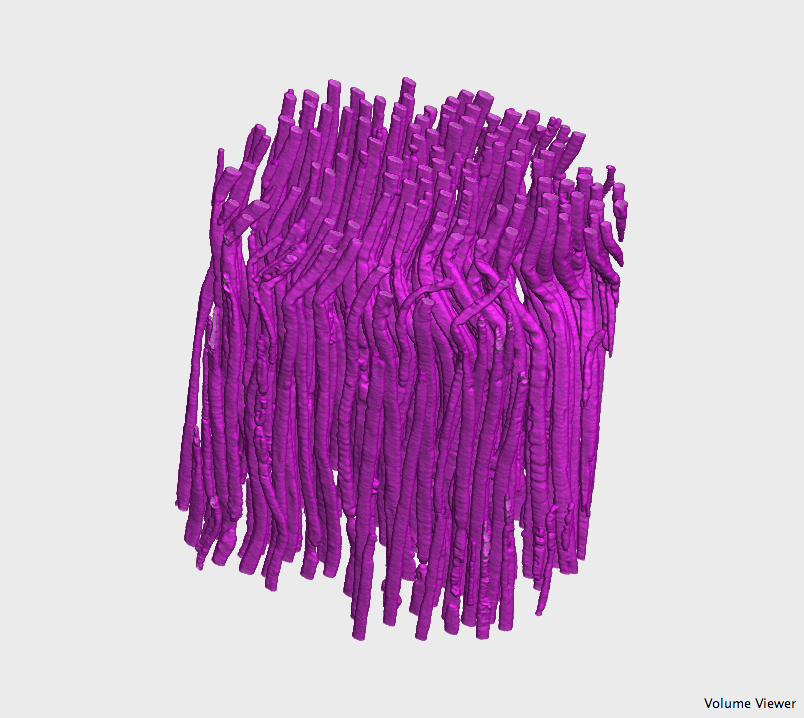}%
  \caption{\label{fig:nerves} Volumetric segmentation of a peripheral nerve. In the \emph{top} row, a slice from the volumetric data with overlayed limited user input and the three-class segmentation dividing the pixels into background (cyan), myelin (purple) and axon (magenta). The \emph{middle} row shows two layers of probability images, corresponding to the myelin class and the axon class. High intensity indicates a high probability of belonging to the class. The \emph{bottom} row shows the 3D visualization of the data obtained by processing a full volume stack and assigning each voxel to the class of the highest probability. This experiment was performed using $M =9$, $K=4000$ and a clustering based on SIFT features.} 
\end{figure}

In Fig.~\ref{fig:nerves} we show a three-class classification of a volumetric X-ray image of a peripheral nerve with myelinated axons appearing as tubular structures. The data originates from a larger study \cite{dahlin2020three} which used our method as initialization for mesh-based segmentation \cite{jeppesen2020sparse}. Using a purely intensity-based approach to pixel classification, it would be difficult to differentiate between the bright background and the bright axons inside the dark myelin. Furthermore, a significant bias field makes it difficult to choose a global threshold. Our approach utilises a very limited user input in just one slice of the volume to differentiate between three classes: background, myelin, and axon. Based on the dictionary probabilities learned from this one slice, our method automatically classifies all other slices yielding a volumetric segmentation. 

\begin{figure}
\centering
  \includegraphics[width=\sub]{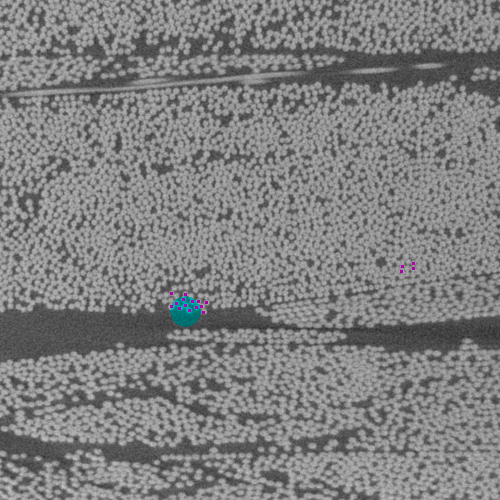}\hspace{\vsep}%
  \includegraphics[width=\sub]{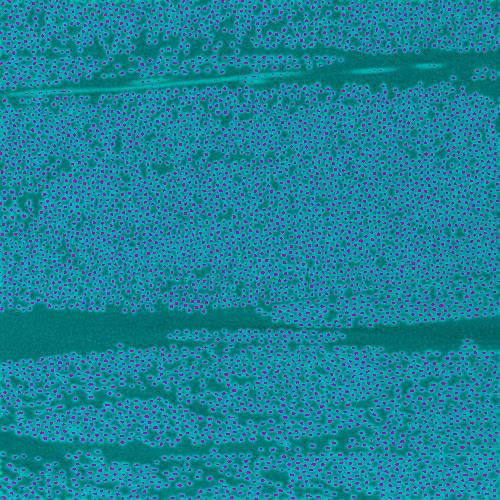}\\\vspace{\vsep}%
  \includegraphics[width=\sub]{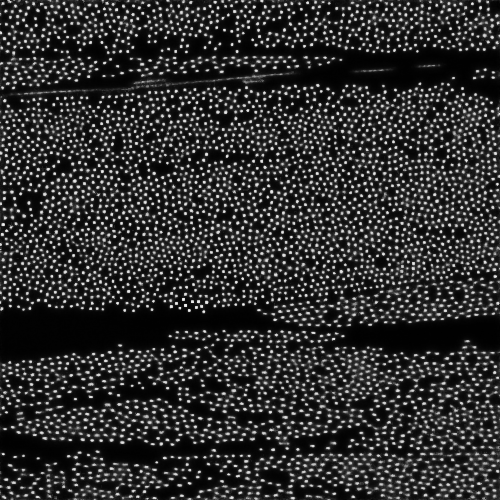}\hspace{\vsep}%
  \includegraphics[width=\sub]{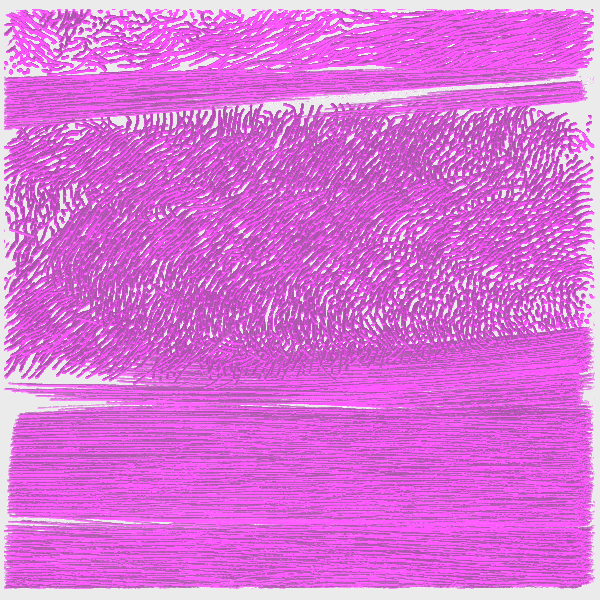}\\\vspace{\vsep}%
  \includegraphics[width=0.3266\linewidth,trim={7cm 7cm 7cm 7cm},clip]{FFigures/settings_labels_displayed.png}\hspace{\vsep}%
  \includegraphics[width=0.3266\linewidth,trim={7cm 7cm 7cm 7cm},clip]{FFigures/P_1.png}\hspace{\vsep}%
  \includegraphics[width=0.3266\linewidth,trim={7cm 7cm 7cm 7cm},clip]{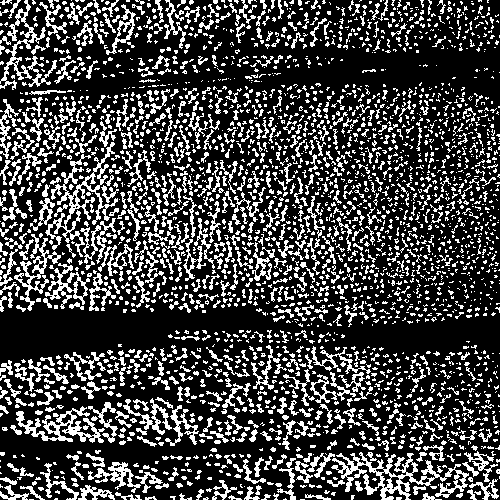}%
  \caption{Volumetric segmentation of fibre composite. In the \emph{top} row a slice with manual input indicating fibre centres (magenta) on a background (cyan) class, together with a resulting two-class segmentation. The \emph{middle} row shows a probability image corresponding to the fibre centres class and the output of processing a full volume stack. From the 3D visualisation it is evident that fibres form clusters of different orientations. The \emph{bottom} row shows a zoom-in on the central part of the image slice, together with the corresponding probability image and (for comparison) a segmentation obtained by directly thresholding the image intensities. Settings used in this experiment are $M=9$ and $K=4000$.\label{fig:fibres2}}
\end{figure}

Fig.~\ref{fig:fibres2} shows an example of segmenting a volumetric image of a fibre composite into two classes: background class and fibre centre class. Using our method, a huge number of individual fibres can be segmented with modest user input. The probability image of a fibre centre class precisely indicates a region for each fibre centre, and can readily be used in postprocessing for obtaining information about the spatial distribution of fibres. In this example we also use the result of single-slice segmentation for batch processing of a whole volume stack, allowing quantification of larger material sample \cite{emerson2018statistical,emerson2018quantifying}. 
For comparison, we also show a result obtained by thresholding the intensity image. This nicely illustrates a challenge in segmenting densely packed fibres, when the image resolution does not suffice to clearly delineate the boundary of every individual fibre. 

\begin{figure}
\centering
  \includegraphics[width=\sub]{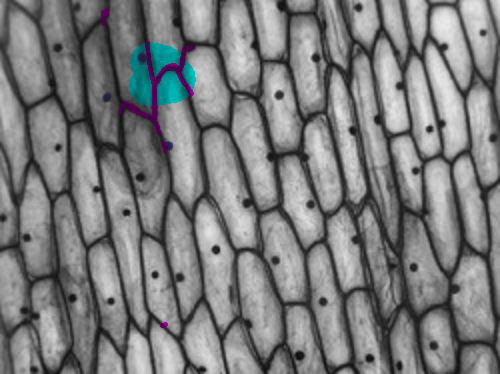}\hspace{\vsep}%
  \includegraphics[width=\sub]{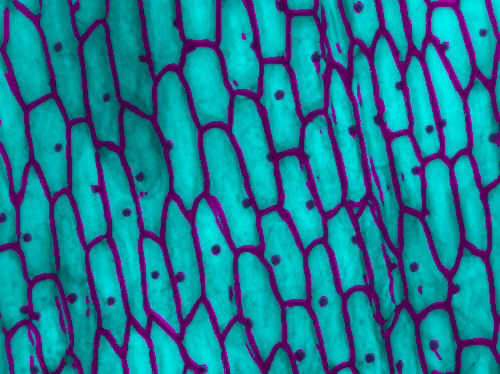}\\\vspace{\vsep}
  \includegraphics[width=\sub]{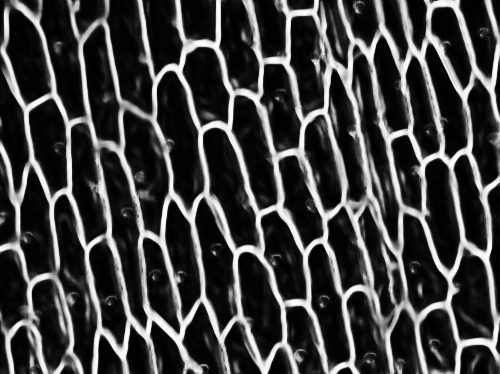}\hspace{\vsep}%
  \includegraphics[width=\sub]{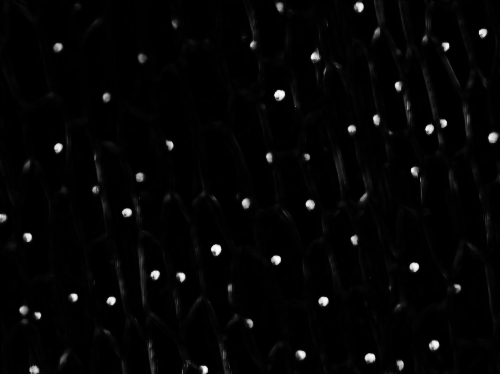}
  \caption{A three-class segmentation of onion cells. In the \emph{top} row an image with manual input and a segmentation into three classes: background (cyan), nucleus (purple) and wall (magenta). In the \emph{bottom} row the probability images for the wall and the nucleus class. Settings used are $M=9$ and $K=4000$.\label{fig:onion}}
\end{figure}

In Fig.~\ref{fig:onion} we show a three-class segmentation of onion cells. Since cell walls and nuclei both appear dark, a purely intensity-based method would not distinguish these two classes -- a task which our method successfully solves with only modest user input.

\begin{figure}
\centering
  \includegraphics[width=\sub]{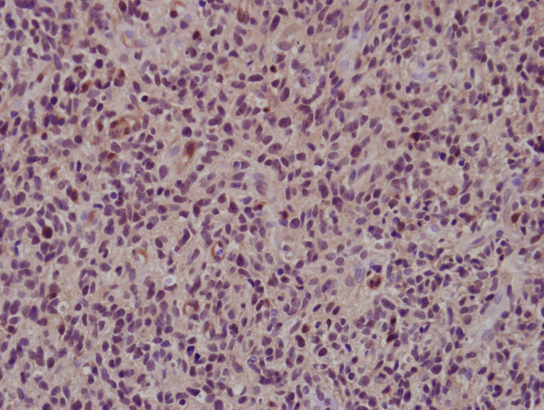}\hspace{\vsep}%
  \includegraphics[width=\sub]{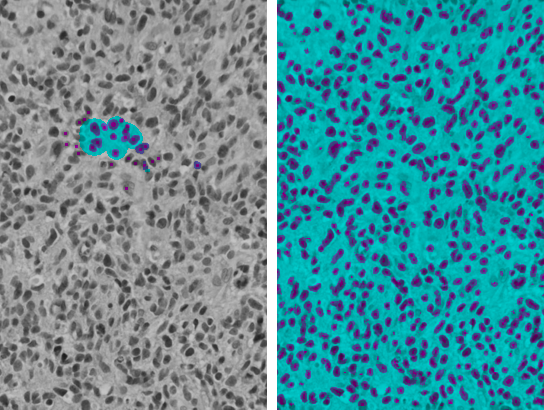}\\\vspace{\vsep}
  \includegraphics[width=\sub]{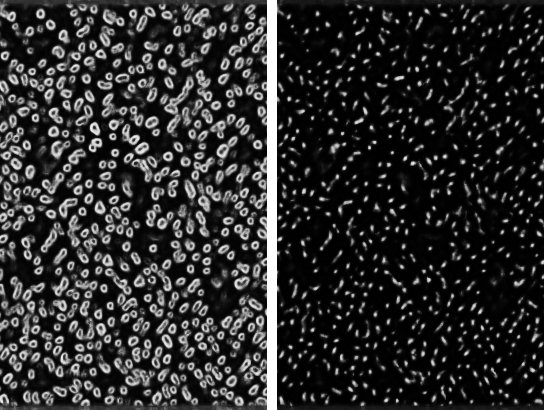}\hspace{\vsep}%
  \includegraphics[width=\sub]{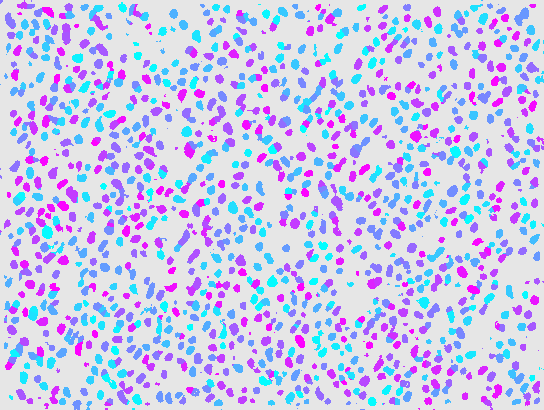}
  \caption{A three-class segmentation of a histopathology image. In the  \emph{top left} an original colour image. In the \emph{top right} the extent of the manual input and a corresponding segmentation into three classes, with a frame cropped to the central part of the image. In the \emph{bottom left} the probability images for the two classes also shown on the central part. In the \emph{bottom right} the final result obtained through additional postprocessing to distinguish individual cells. Settings used are $M=5$ and $K=4000$.\label{fig:histo}}
\end{figure}

In Fig.~\ref{fig:histo} we show the use of our method for counting cells in a stained microscopy image, similar to \cite{kaarsnas2011learning}. Unlike other examples, this is a colour (RGB) image. To utilise colour information, the features extracted from every image patch contain three colour channels concatenated into a single feature vector. Since the final goal is to count and measure the cells, we postprocess the probability images. This is done by computing the local maxima of the centre-class probability image to obtain individual cell segmentation.

\section{Conclusion}
\label{sec:conclusion}

We propose a method for interactive labeling of image pixels. Instrumental for our method are transformations which propagate the information from the image grid to a dictionary, and back to the image. The transformations are constructed such that the propagation is strong between image pixels with a similar appearance. We present an algorithm for building a matrix representation of those transformations, allowing real-time processing. We demonstrate how the propagation of user-provided labelling can be used for interactive image segmentation. Furthermore, a segmentation of one image allows for subsequent automatic processing of similar images. With only modest user input, our method can yield good results when segmenting patterned images. We find this extremely useful for many tasks in microscopy for materials and life sciences.


\paragraph{Acknowledgment} This work is supported by The Center for Quantification of Imaging Data from MAX IV (QIM) funded by The Capital Region of Denmark.

{\small
\bibliographystyle{ieee_fullname}
\bibliography{texture_gui_bib}
}

\end{document}